\documentclass{article}
\usepackage{spconf,amsmath,graphicx,hyperref}
\usepackage{xcolor}
\usepackage{tabularx}
\usepackage{booktabs}

\title{Advancing Speech Understanding in Speech-Aware Language Models with GRPO}
\name{Avishai Elmakies* \thanks{*Work done during internship at IBM Research}, Hagai Aronowitz, Nimrod Shabtay, Eli Schwartz, Ron Hoory, Avihu Dekel}
\address{IBM Research}
\begin{document}
%
\maketitle

\begin{abstract}
In this paper, we introduce a Group Relative Policy
Optimization (GRPO)-based method for training Speech-Aware Large Language Models (SALLMs) on open-format speech understanding tasks, such as Spoken Question Answering and Automatic Speech Translation. SALLMs have proven highly effective for speech understanding tasks. GRPO has recently gained traction for its efficiency in training LLMs, and prior work has explored its application to SALLMs, primarily in multiple-choice tasks. Building on this, we focus on open-format tasks that better reflect the generative abilities of the models. Our approach leverages GRPO with BLEU as the reward signal to optimize SALLMs, and we demonstrate empirically that it surpasses standard SFT across several key metrics. Finally, we explore the potential of incorporating off-policy samples within GRPO for these tasks, highlighting avenues for further improvement and further research.

\end{abstract}

\newcommand{\newpara}[1]{\vspace{0.075cm} \noindent {#1}}

\newtoggle{release}

\togglefalse{release}

\iftoggle{release}{
    \newcommand{\av}[1]{}
    \newcommand{\avihu}[1]{}
    \newcommand{\nimrod}[1]{}
}
{
    \newcommand{\av}[1]{\textcolor{green}{av: #1}}
    \newcommand{\avihu}[1]{\textcolor{red}{Avihu: #1}}
    \newcommand{\nimrod}[1]{\textcolor{blue}{nimrod: #1}}
    
}
\section{introduction}

Reinforcement Learning (RL) has gained significant traction, due to its effectiveness in improving the reasoning capabilities of LLMs and VLMs \cite{shao2024deepseekmath, huang2025vision}. 
Inspired by the advancements in Multi-Modal RL, we aim to improve Speech-Aware Large Language Models (SALLMs) \cite{arora2025landscape}.
SALLMs process spoken language by taking combined speech-and-text inputs and generating text outputs. They are widely used for tasks like Automatic Speech Recognition (ASR), Automatic Speech Translation (AST), and Spoken Question Answering (SQA) \cite{saon2025granite, abouelenin2025phi}.

Recent works have applied RL to SALLMs, typically for SQA, yet these approaches often rely on binary rewards, which constrain the model's ability to provide open-ended answers \cite{rouditchenko2025omni, li2025reinforcement}. 
Other works use unsupervised methods for reward calculation \cite{wang2025self}, but these often yield subpar results compared to standard supervised fine-tuning (SFT).

\textbf{In this work:} we propose an RL approach with verifiable rewards (RLVR) to improve SALLMs. Our method leverages the GRPO algorithm \cite{shao2024deepseekmath} and uses the BLEU metric (or other metrics) \cite{papineni2002bleu} as a reward signal. We evaluate our approach on speech understanding tasks -- SQA and AST. Our approach demonstrates strong performance on open-ended responses, yielding superior results to standard supervised fine-tuning (SFT) as well as the baseline model. We show this empirically using standardized metrics such as BLEU \cite{papineni2002bleu}, ROUGE \cite{lin2004rouge}, METEOR \cite{banerjee2005meteor}, and BERTScore \cite{zhang2019bertscore}. Our contributions are summarized as follows:
\begin{itemize}
    \setlength\itemsep{-0.2em}
    \item We propose to train SALLMs by integrating GRPO with various reward functions.
    \item We empirically demonstrate that our approach outperforms standard SFT on SQA and AST across multiple relevant metrics. 
    \item We study the effectiveness of using off-policy samples as part of our proposed method.  
\end{itemize}

{\fontsize{10}{12}\selectfont
\begin{table*}[ht] 
\caption{Results on LibriSQA (SQA). SOTA results are taken from LibriSQA paper \cite{zhao2024librisqa}.}  
\label{results_sqa}
\begin{tabular}{@{}lcccccc@{}}
\toprule
& \textbf{BLEU} $(\uparrow)$ &
\textbf{BERTScore} $(\uparrow)$ &
\textbf{ROUGE-1} $(\uparrow)$ &
\textbf{ROUGE-2} $(\uparrow)$ &
\textbf{ROUGE-L} $(\uparrow)$ &
\textbf{METEOR} $(\uparrow)$ \\ \midrule
SOTA            & 33.78          & 93.07          & 65.38          & 50.19          & 62.09          & -              \\ \midrule
Granite Speech 2B & 27.74          & 91.17          & 56.66          & 40.25          & 51.26          & 53.01          \\
+ SFT             & 40.88          & 94.15          & 65.13          & 49.07          & 61.50          & 64.64          \\
+ GRPO            & \textbf{44.90} & \textbf{94.45} & \textbf{68.56} & \textbf{53.35} & \textbf{64.88} & \textbf{68.48} \\ \midrule
Granite Speech 8B & 17.85          & 90.25          & 49.58          & 34.31          & 43.05          & 53.19          \\
+ SFT             & 42.34          & 94.49          & 67.05          & 51.54          & 63.76          & 65.99          \\
+ GRPO            & \textbf{46.40} & \textbf{94.76} & \textbf{69.57} & \textbf{57.49} & \textbf{66.16} & \textbf{69.61} \\ \bottomrule
\end{tabular}
\label{tab:sqa_results}
\end{table*}
}

{\fontsize{10}{12}\selectfont
\begin{table*}[t] 
\caption{Results on CoVoST2 English to German (AST). SOTA results are taken from Phi-4-mini paper (Table 5) ~\cite{abouelenin2025phi}. }
\label{results_ast}
\begin{tabular}{@{}lcccccc@{}}
\toprule
& \textbf{BLEU} $(\uparrow)$ &
\textbf{BERTScore} $(\uparrow)$ &
\textbf{ROUGE-1} $(\uparrow)$ &
\textbf{ROUGE-2} $(\uparrow)$ &
\textbf{ROUGE-L} $(\uparrow)$ &
\textbf{METEOR} $(\uparrow)$ \\ \midrule
SOTA            & 37.16          & -          & -          & -          & -          & -              \\ \midrule
Granite Speech 2B & 29.06          & 86.04          & 57.25          & 35.19          & 54.09          & 55.03          \\
+ SFT             & 30.50          & 86.40          & 58.53          & 36.75          & 55.21          & 56.18          \\
+ GRPO            & \textbf{31.47} & \textbf{86.90} & \textbf{59.99} & \textbf{37.88} & \textbf{56.75} & \textbf{57.48} \\ \midrule
Granite Speech 8B & 32.48          & 87.26          & 60.48          & 38.78          & 57.17          & 58.24          \\
+ SFT             & 31.62          & 86.76          & 59.66          & 37.91          & 56.35          & 57.35          \\
+ GRPO            & \textbf{35.08} & \textbf{87.64} & \textbf{62.90} & \textbf{41.40} & \textbf{59.64} & \textbf{60.40} \\ \bottomrule
\end{tabular}
\label{tab:ast_results}
\end{table*}
}

\section{Related Work}

SALLMs have gained recent attention, driven by their potential in speech understanding tasks such as ASR, AST, and SQA. These models leverage a speech encoder, a language model, and train a lightweight projector to align the two modalities.
Leading models such as \cite{saon2025granite, abouelenin2025phi} are trained on extensive speech datasets to improve their capabilities. Unlike Speech Language Models (SLMs) \cite{lakhotia2021generative, maimon2025slamming} which take speech as input and output speech, SALLMs take speech as input but output only text, which makes them more appropriate for chat-oriented tasks.

Reinforcement Learning methods have been used to improve the capabilities of LLMs. 
Proximal Policy Optimization (PPO) \cite{schulman2017proximal} was used to help LLMs learn from human feedback \cite{ouyang2022training}, but requires training a reward model and has high memory requirements. 
Direct Preference Optimization (DPO) demonstrates that when preference response pairs are available, LLMs can be trained without reward \cite{rafailov2023direct}.

GRPO is an on-policy Reinforcement Learning algorithm, learning from its own generated data.
Instead of relying on a value model, it updates the policy and obtains advantage estimates using an empirical normalization of the grouped samples (see Eq.~\ref{eq:advantage}). GRPO has been shown to produce SOTA results in reasoning tasks such as math and coding \cite{shao2024deepseekmath}.

While GRPO has also been used to improve SQA in SALLMs \cite{rouditchenko2025omni, li2025reinforcement}, these studies focused on multi-choice datasets and benchmarks \cite{sakshi2024mmau, wang2025mmsu,rouditchenko2025omni, li2025reinforcement}. 
While easy to evaluate, better performance on those datasets may not reflect the generative capabilities of the evaluated model in tasks like open-ended QA, a key requirement in chat-oriented applications. 

A recent approach incorporated off-policy samples into the sample group \cite{yan2025learning}, allowing the model to learn from both on-policy samples and off-policy samples (which were acquired using a different approach/policy). This approach, known as mixed-policy, can be particularly beneficial when high-quality off-policy data is available.
Most work on this approach explores text-based models with reasoning tasks using off-policy samples from closed models, such as ChatGPT. 
In this work, we explore the effect of adding the ground truth to our sample group in GRPO when training SALLMs. 

Metrics like BLEU \cite{papineni2002bleu}, BERTScore \cite{zhang2019bertscore}, ROUGE \cite{lin2004rouge}, and METEOR \cite{banerjee2005meteor} are widely used to evaluate the similarity of a candidate text response to some text reference.
While commonly used as a metric for tasks such as machine translation, these metrics have also been adopted as reward functions in RL-based approaches to improve machine translation \cite{choshen2019weaknesses}.
More recently, BLEU has also been used as a reward alongside GRPO to improve text-only LLMs on instruction following tasks \cite{chang2025bleuberi}. 
We investigate using BLEU and other metrics, computed between the ground-truth and generated text, as a reward for speech-related tasks in SALLMs.

\begin{table*}[ht]
\centering
\caption{Comparing different reward functions when training Granite Speech 2B on LibriSQA}
{\fontsize{10}{12}\selectfont
\begin{tabular}{@{}llllllll@{}}
\toprule
Reward  & BLEU  & BERTScore & ROUGE-1 & ROUGE-2 & ROUGE-L & METEOR & AVG   \\ \midrule
BLEU    & \textbf{44.9}  & \textbf{94.45}     & 68.56   & 53.35   & 64.88   & 68.48  & \textbf{65.77} \\
ROUGE-1 & 38.81 & 93.54     & \textbf{68.87}   & 53.45   & 64.76   & 60.65  & 63.35 \\
ROUGE-2 & 37.82 & 93.52     & 68.59   & \textbf{54.15}   & 65.27   & 58.87  & 63.04 \\
ROUGE-L & 37.95 & 93.56     & 68.68   & 53.84   & \textbf{65.44}   & 59.27  & 63.12 \\
METEOR  & 37.69 & 94.04     & 66.99   & 51.74   & 62.63   & \textbf{70.25}  & 63.89 \\ \bottomrule
\end{tabular}
}
\label{table:reward_ablation}
\end{table*}
\section{Method}

\subsection{GRPO}
\label{subsec:rl}
In this work, we propose to use GRPO as our training algorithm \cite{shao2024deepseekmath}, which samples different responses from the model, calculates their reward, and uses the GRPO loss to increase the likelihood of high-reward responses.
For each prompt $q\sim\mathcal{D}$, we sample $G$ times from our policy $\pi_\theta$. For each sample $\textbf{G}=\{o_1,\dots,o_G\}$ we calculate a reward $\textbf{R}=\{r_1,\dots,r_G\}$ then use the rewards to estimate the advantages:
\begin{equation}
   \hat{A}_i = \frac{r_i - mean(\textbf{R})}{std(\textbf{R})}
\label{eq:advantage}
\end{equation}
Using the above rewards, we can optimize a loss. Specifically, we use the loss introduced in DAPO \cite{yu2025dapo}, a slight variation of the original GRPO loss.

\begin{equation}
    l_{i,t} =  min(s_{i,t}(\theta)\hat{A}_{i},clip[s_{i,t}(\theta), 1 - \epsilon, 1 + \epsilon]\hat{A}_{i})
\end{equation}

where we have $s_{i,t}(\theta) = \frac{\pi_\theta(o_{i,t}|p,o_{i,<t})}{\pi_{old}(o_{i,t}|p,o_{i,<t})}$ is the importance sampling weight. This gives us the final loss:

\begin{align}
    \mathcal{L}_{DAPO}(\theta) = -\frac{1}{|\textbf{G}|}\sum_{i=1}^G \sum_{t=1}^{|o_i|} (l_{i,t} - \beta D_{KL}[\pi_\theta || \pi_{ref}])
\end{align}

where $|\textbf{G}| = \sum_{i=1}^G |o_i|$ i.e., the number of tokens in the completions generated by the policy.

\subsection{Mixed-Policy GRPO}

Yan et al. \cite{yan2025learning} define $\textbf{G}$ as a group of both on-policy samples and off-policy samples, estimating rewards as in Eq. \ref{eq:advantage}. 
They define the off-policy importance sampling weight:
\begin{align}
    \hat{s}_{j,t}=\frac{\pi_\theta(o_{j,t}|p,o_{j,<t})}{\pi_{\phi}(o_{j,t}|p,o_{j,<t})} \label{eq:off_policy_importance_sample}
\end{align}

where $\pi_\phi$ is the policy that generated the off-policy samples, and suggest setting $\pi_\phi=1$ and disabling clipping for those samples.
We explore incorporating the ground truth reference as an off-policy sample in the group to help steer the policy toward high-quality generations, especially early in training. We will denote this method as MP-GRPO.
Since we don't know $\pi_\phi$, we follow their recommendations and set $\pi_\phi=1$ and disable clipping. We use a single off-policy sample in the group, the reference.We note this could be extended to multiple references, but we leave this for future work. 

\subsection{Reward functions}
Given some text predicted by our policy, we calculate its reward by computing its BLEU score to a ground-truth reference. The resulting score is within $[0,1]$, indicating the text similarity. In our case we will use BLEU to compare a reference and samples from our policy. BLEU is highly suitable for generative open-ended tasks, with several possible answers. 
We will note that there are many possible rewards similar to BLEU which could be applicable for our setup, we compare BLEU to other possible rewards in Section~\ref{sub:other_rewards}. Neural-based rewards, such as BERT-score \cite{zhang2019bertscore}, or possibly a combination of rewards may also be considered. Due to computational constraints, we focus on BLEU and similar rewards, and leave other possibilities for future work.


\section{Experimental Setup}

\subsection{Tasks and datasets}
In this work, we focus on SQA and AST. We also performed some preliminary experiments with ASR and obtained only minor improvements. 
Unlike ASR, which has a unique valid response for each utterance, tasks like AST and open-ended SQA can have multiple valid outputs. This characteristic makes them particularly well-suited for our sampling-based approach, where we anticipate greater performance gains. We believe that further work needs to be done on the abilities of RL/GRPO in tasks such as ASR.

For open-ended SQA we use LibriSQA part I \cite{zhao2024librisqa} as our dataset. 
This dataset is based on LibriSpeech \cite{panayotov2015librispeech} and contains spoken utterances taken from audio-books as well as questions and open answers created by an LLM. 
This dataset contains $107K$ training samples. We use $20\%$ for validation and the test set has about $2500$ samples. We use multiple possible prompts for this task, e.g. \texttt{"Listen to the audio <|audio|> and answer the following question:<|question|>"}. 

For AST we use CoVoST2 \cite{wang2021covost}, focusing on English to German translation. This dataset contains spoken utterances in English and text translations in German. 
We prompt the model to translate the speech utterance to German directly, without using ASR. This dataset contains about $220K/12K/15K$ in the train/validation/test splits. We use multiple possible prompts, e.g. \texttt{"<|audio|> listen to the speech and translate it into German."}

\subsection{Models configuration}

It is important to note that SFT and GRPO are two very different approaches. This means that a fully comparable setup is difficult to achieve. Therefore, we report test set performance using the best models selected based on validation results. We believe that this is the fairest comparison between the approaches.

Most of our experiments were done on Granite Speech $2B$~\cite{saon2025granite}, which is based on Granite $3.3$~\cite{granite2024granite}, a CTC Speech encoder and a Window Q-Former projector. Some are based on Granite Speech $8B$~\cite{saon2025granite}, to test the scalability of our approach. We note that Granite Speech was not trained on SQA. These models were trained on AST, but not on CoVoST2. 

We optimized models with AdamW \cite{loshchilov2017fixing} and performed hyperparameter searches for both SFT and GRPO, selecting the best validation model per task. SFT covered learning rates $[10^{-6},5\cdot10^{-5}]$, epochs $[1,10]$, batch sizes $[12,48]$, and warm-up $[0,0.15]$. GRPO used similar ranges plus $\beta \in [0,0.04]$, group size $[4,12]$, prompt length 256, and max completion length 200.
GRPO used temperature $=1$ during training. All models used top-p $=0.9$, temperature $=0.9$ for evaluation. 
With GRPO, a group size of $G=8$ and $\beta=0.02$ yielded stable training. Setting $\beta=0$ led to divergence, especially on larger datasets, while higher $\beta$ reduced reward performance.
When following the mixed-policy scheme, we used $G-1$ on-policy samples and $1$ off-policy.
Training used 4 H100 GPUs; GRPO on Granite Speech $2B$ took up to 24h, compared to far lower SFT cost. Comparable compute budgets for SFT showed minimal gains.

\begin{figure}[t]
  \centering
  \centerline{\includegraphics[width=8.5cm]{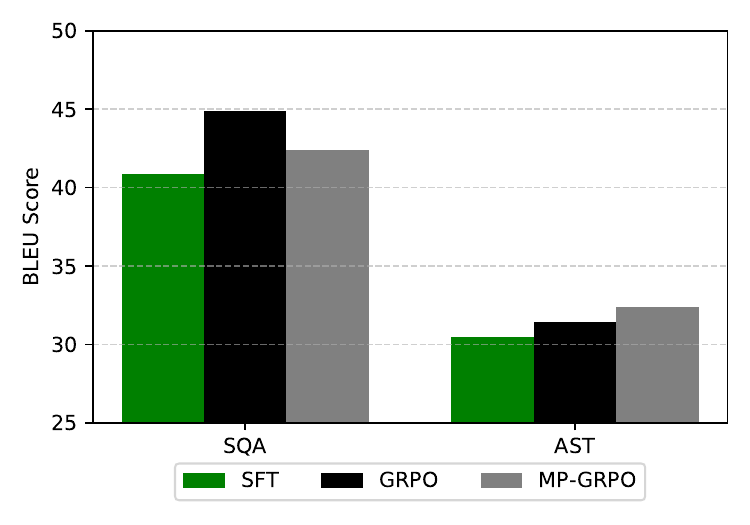}}
\caption{SFT, GRPO and MP-GRPO on SQA and AST}
\label{fig:grpo_vs_mpgrpo}
\end{figure}

\section{results}

In this section, we report the results of our experiment.  We compare the results of our method to other approaches using BERTScore-F1, BLEU, ROUGE-1/2/L and METEOR, metrics suited to comparing the model's generated outputs.

\subsection{Spoken question answering}
Table~\ref{tab:sqa_results} presents SQA results on the LibriSQA dataset. For both model sizes, the baseline scores are lower than the SOTA results reported in the LibriSQA paper \cite{zhao2024librisqa}.
Using either SFT or GRPO improves the results by a large margin, surpassing the SOTA results. Finally, we observe that training with our GRPO approach, outperforms training using SFT. 
Compared to the base and SFT models, Granite Speech $2B$ yields BLEU improvements of $61.8\%$ and $9.8\%$ respectively, whereas the $8B$ model achieves $151\%$ and $6\%$ respectively.

\subsection{Automatic speech translation}

Results for the task of AST on CoVoST2 English to German are presented in Table~\ref{tab:ast_results}. 
Granite Speech $2B$ sees improvements from both SFT and GRPO, with GRPO achieving the strongest results ($+8.2\%$ BLEU over base, $+3.2\%$ over SFT). In contrast, Granite Speech $8B$ shows degraded performance with SFT, while GRPO continues to provide gains ($+8\%$ over base, $+10.9\%$ over SFT).



\subsection{Mixed-Policy GRPO}

We examine the effect of Mixed-Policy GRPO compared to our standard GRPO approach. We report the results for both SQA and AST on Granite Speech 2B. We present results for BLEU in Fig.~\ref{fig:grpo_vs_mpgrpo}, and note that other metrics show similar trends. We observe that MP-GRPO improves the BLEU score on AST compared to GRPO, suggesting that for AST, adding the reference to the training with GRPO has the potential to improve results. In contrast, when we compare MP-GRPO to GRPO on SQA we observe that the performance degrades when using the reference with GRPO.
This difference may be related to the fact that the base model was not trained on SQA but was trained on AST. With SQA, the model had much more to learn, which could have made the off-policy training less stable, resulting in degraded performance. With AST, the model had less to learn, and the off-policy samples were used as anchors that the model could learn from.

\subsection{Different rewards options} \label{sub:other_rewards}

Different text-comparison metrics could be computed and used as a reward. To show that BLEU is the right choice for our approach we trained Granite Speech 2B on LibriSQA, using reward functions including BLEU, ROUGE-1/2/L and METEOR. We tested the overall impact of the reward function on different performance metrics.
Results in Table~\ref{table:reward_ablation} first show that optimizing a given metric yields the best results on that metric, usually at the cost of other metrics, especially BLEU. 
Among all evaluated metrics, the BLEU score yielded the highest average performance, indicating its overall superiority compared to the other methods considered.

\section{conclusion}

In this work, we propose a method for training Speech Aware Large Language Models. We use GRPO and BLEU to improve the capabilities of SALLMs on open-ended tasks, specifically open-ended Speech Question Answering and Automatic Speech Translation. Unlike multiple-choice tasks, open-format tasks provide a more effective means of assessing the generative abilities of SALLMs.

We show that despite its simplicity, our approach is highly effective. We observe improvements across all metrics; results match or outperform those obtained using the SFT approach.
We also show that this approach scales to larger models, showing improvements for models of sizes $2B$ and $8B$.

Finally, we experimented with mixed-policy GRPO, which leverages both on-policy samples and off-policy samples to train our model. While this approach shows promise, further investigation is required, and we encourage the research community to pursue it.
We hope this work will inspire the community to test the boundaries of both on-policy algorithms, off-policy algorithms, and even mixed-policy algorithms on different Speech Understanding tasks.

\newpage

\bibliographystyle{IEEEbib}
\bibliography{refs}

\end{document}